\pdfoutput=1
\documentclass[10pt]{article}
\usepackage{times}

\usepackage{graphicx}
\usepackage{hyperref}
\usepackage{caption}
\usepackage{lineno}

\usepackage{xcolor}
\definecolor{defaultcolor}{gray}{0.9}
\newlength\savewidth\newcommand\shline{\noalign{\global\savewidth\arrayrulewidth
  \global\arrayrulewidth 1pt}\hline\noalign{\global\arrayrulewidth\savewidth}}
\newcommand{\tablestyle}[2]{\setlength{\tabcolsep}{#1}\renewcommand{\arraystretch}{#2}\centering\footnotesize}

\usepackage{amsmath,amsfonts,bm}

\definecolor{defaultcolor}{gray}{0.9}

\def\eqref#1{equation~\ref{#1}}

\def\1{\bm{1}}

\def\mI{{\bm{I}}}

\def\mT{{\bm{T}}}

\DeclareMathAlphabet{\mathsfit}{\encodingdefault}{\sfdefault}{m}{sl}
\SetMathAlphabet{\mathsfit}{bold}{\encodingdefault}{\sfdefault}{bx}{n}

\usepackage[symbol]{footmisc}
\usepackage[noblocks]{authblk}

\usepackage{booktabs}
\usepackage{graphicx}
\usepackage{multirow}
\usepackage{color, colortbl}
\usepackage{caption}
\usepackage{subcaption}
\usepackage{wrapfig}
\usepackage{makecell}
\usepackage{amsfonts}
\usepackage{amsmath}
\usepackage{booktabs}
\usepackage{xspace}
\usepackage{array}
\usepackage{siunitx}
\usepackage{enumitem}
\usepackage[most]{tcolorbox}
\usepackage[square, numbers, sort&compress]{natbib}
\usepackage[margin=1in]{geometry}
\usepackage[normalem]{ulem}
\usepackage{fancyhdr}
\usepackage{soul}

\newcolumntype{F}[1]{%
    >{\raggedright\arraybackslash\hspace{0pt}}p{#1}}%
\newcolumntype{T}[1]{%
    >{\centering\arraybackslash\hspace{0pt}}p{#1}}%

\usepackage[margin=1in]{geometry}
\usepackage[symbol]{footmisc}

\newcommand{\ourmodel}{BiomedCLIP}
\newcommand{\ourdata}{PMC-15M}
\newcommand{\ourdatalarge}{PMC-Fine-Grained-46M}
\newcommand{\eat}[1]{\ignorespaces}

\usepackage[capitalize]{cleveref}
\crefformat{section}{\S#2#1#3}
\Crefname{figure}{\textbf{Figure}}{}
\Crefname{table}{\textbf{Supplementary Table}}{}
\Crefname{algorithm}{Algorithm}{}
\Crefname{algocf}{Algorithm}{}
\Crefname{equation}{Equation}{}
\crefname{appendix}{Appendix}{}
\title{BiomedCLIP: a multimodal biomedical foundation model pretrained from fifteen million scientific image-text pairs}

\author{
Sheng Zhang$^{1,*}$, Yanbo Xu$^{1,*}$, Naoto Usuyama$^{1,*}$, Hanwen Xu$^{2,*}$, Jaspreet Bagga$^{1}$, \\
Robert Tinn$^{1}$, Sam Preston$^{1}$, Rajesh Rao$^{1}$, Mu Wei$^{1}$, Naveen Valluri$^{1}$, Cliff Wong$^{1}$, \\
Andrea Tupini$^{1}$, Yu Wang$^{1}$, Matt Mazzola$^{1}$, Swadheen Shukla$^{1}$, Lars Liden$^{1}$, Jianfeng Gao$^{1}$, \\
Angela Crabtree$^{3}$, Brian Piening$^{3}$, Carlo Bifulco$^{3}$, Matthew P. Lungren$^{1}$, Tristan Naumann$^{1}$, Sheng Wang$^{2,\ddag}$, and Hoifung Poon$^{1,\ddag}$ \\
    \vspace{1em}
    $^1$Microsoft Research, Redmond, WA \\
    $^2$Paul G. Allen School of Computer Science and Engineering, University of Washington, Seattle, WA\\
    $^3$Providence Genomics, Portland, OR
}
\footnotetext{* Equal contribution}
\footnotetext{$\ddag$ Corresponding authors. Emails: swang@cs.washington.edu, hoifung@microsoft.com}

\date{}
\begin{document}
\maketitle
\begin{abstract}
\noindent Biomedical data is inherently multimodal, comprising physical measurements and natural language narratives. A generalist biomedical AI model needs to simultaneously process different modalities of data, including text and images. Therefore, training an effective generalist biomedical model requires high-quality multimodal data, such as parallel image-text pairs. Here, we present \ourdata{}, a novel dataset that is two orders of magnitude larger than existing biomedical multimodal datasets such as MIMIC-CXR, and spans a diverse range of biomedical image types. \ourdata{} contains 15 million biomedical image-text pairs collected from 4.4 million scientific articles. Based on \ourdata{}, we have pretrained \ourmodel{}, a multimodal foundation model, with domain-specific adaptations tailored to biomedical vision-language processing. We conducted extensive experiments and ablation studies on standard biomedical imaging tasks from retrieval to classification to visual question-answering (VQA). BiomedCLIP achieved new state-of-the-art results in a wide range of standard datasets, including retrieval, image classification, and visual question answering, substantially outperforming prior approaches. Intriguingly, by large-scale pretraining on diverse biomedical image types, BiomedCLIP even outperforms state-of-the-art radiology-specific models such as BioViL in radiology-specific tasks such as RSNA pneumonia detection. We further illustrated how BiomedCLIP could promote privacy-preserving analysis of proprietary data by using \ourdata{} as a proxy. In summary, \ourmodel{} is a fully open-access foundation model that achieves state-of-the-art performance on various biomedical tasks, paving the way for transformative multimodal biomedical discovery and applications. We release our models at \href{https://aka.ms/biomedclip}{aka.ms/biomedclip} to facilitate future research in multimodal biomedical AI. 

\end{abstract}


\newpage
\section*{Introduction}
Biomedical data is inherently multimodal, comprising both physical measurements and natural-language narratives \citep{moor2023foundation,tu2023towards}. However, only a fraction of such data is interpreted and made available in structured forms, as manual curation is costly and hard to scale. Multimodal learning can alleviate the curation bottleneck by leveraging self-supervision from cross-modal correspondence. It also helps uncover predictive signals that may remain latent in standalone modalities through multimodal fusion. As a result, biomedical vision-language foundation models \citep{heiliger2022beyond,huang2020fusion,ikezogwo2023quilt,zhang2023biomedgpt,huang2023visual,chen2023medblip,xu2024whole,moor2023med} have been developed to collectively model biomedical images (e.g., digital pathology and radiology images) and biomedical text (e.g., scientific papers and clinical notes). There are two key advantages of biomedical vision-language foundation models. First, by pretraining on large-scale data using self-supervised learning, these models excel in low-resource settings, where only a small amount of fine-tuning data is accessible, across a diverse range of downstream applications, thereby substantially decreasing the cost associated with data annotation. Second, by pretraining on high-quality parallel image-text data, these models can solve tasks that require a comprehensive understanding of both image and text data, such as image description generation and visual question-answering.

Since both advantages rely on high-quality biomedical image text data, obtaining such data is crucial to building an accurate and generalizable model. In contrast to the pretraining data used by general-domain vision-language models \citep{radford2021learning, ramesh2021zero, rombach2022high}, the pretraining data used by existing biomedical vision-language models present three major limitations. First, many pretraining data are private data, resulting in the inaccessibility of many biomedical foundation models. Second, existing parallel image-text datasets in the biomedical domain are relatively small, ranging from 7k to 377k pairs (e.g., 377,110 pairs for MIMIC-CXR \citep{johnson2019mimic}, 224,316 pairs for CheXpert \citep{irvin2019chexpert}, 7,562 pairs for ARCH \citep{gamper2021multiple}, and 87,952 pairs for ROCO \citep{pelka2018radiology}). Finally, existing datasets are generally lacking in diversity and most of them focus on chest X-ray, thus limiting their generalizability to other biomedical image types. Prior studies in general domains have demonstrated the advantage of pretraining on diverse, expansive datasets \citep{radford2021learning}, with increasing efforts 
on mining web images and captions \citep{sharma-etal-2018-conceptual,changpinyo2021conceptual,srinivasan2021wit,schuhmann2022laion}, yet biomedical image-text pairs are relatively scarce in such web resources and the quality varies.

In this paper, we develop \ourdata{}, a large biomedical dataset with high-quality parallel image-text pairs extracted from scientific publications in PubMed Central (PMC)\footnote{\url{https://www.ncbi.nlm.nih.gov/pmc/}}. Comprising 15 million biomedical image-text pairs, \ourdata{} aims to address the three limitations in prior biomedical vision-language models. First, \ourdata{} is a public dataset collected from scientific papers, which are devoid of privacy issues. Foundation models trained on \ourdata{} are thus openly accessible. Second, \ourdata{} is two orders of magnitude larger than prior large datasets such as MIMIC-CXR. 
Finally, \ourdata{} covers essentially any category of interest to biomedical research, spanning thirty major biomedical image types, offering a diverse and representative dataset for biomedical research and clinical practice.

Based on \ourdata{}, we have pretrained \ourmodel{}, a state-of-the-art biomedical vision-language foundation model that excels in a wide range of downstream applications such as cross-modal retrieval, zero-shot image classification, and medical visual question answering. \ourmodel{} outperformed the general-domain CLIP model \citep{radford2021learning} by a wide margin across the board, confirming the importance of exploring domain-specific models for biomedicine. 
By pretraining on much larger and more diverse data, \ourmodel{} also substantially outperformed prior state-of-the-art biomedical vision-language models such as PubMedCLIP and MedCLIP, especially in low-resource settings. Surprisingly, BiomedCLIP even outperformed radiology-specific state-of-the-art models such as BioViL~\citep{boecking2022making} on radiology-specific tasks such as RSNA pneumonia detection~\citep{shih2019augmenting}, highlighting the potential of positive transfer in pretraining from diverse image categories. Finally, we explore the possibility of using \ourdata{} as a proxy for analyzing proprietary data using external models. To facilitate future biomedical multimodal research, we will release our \ourmodel{} models and scripts to reproduce \ourdata{} at \href{https://aka.ms/biomedclip}{aka.ms/biomedclip} upon publication.

\eat{
\ourmodel{} also outperformed PubMedCLIP and MedCLIP, which are both fine-tuned to radiology datasets, demonstrating the importance of pretraining on diverse biomedical images. Moreover, with large-scale pretraining across diverse biomedical image types, BiomedCLIP even outperformed radiology-specific state-of-the-art models such as BioViL~\citep{boecking2022making} on radiology-specific tasks such as RSNA pneumonia detection~\citep{shih2019augmenting}. To facilitate future research in biomedical image and text analysis, we release our \ourmodel{} models and scripts to reproduce \ourdata{} at \href{https://aka.ms/biomedclip}{aka.ms/biomedclip}.
}

\section*{Results}
\subsection*{Overview of dataset, model and benchmark}
\paragraph{Dataset} We have created \ourdata{}, a large parallel image-text dataset from scientific papers collected from PubMed Central (PMC). See \textbf{Fig. \ref{fig:overview}A, B}. PMC is a comprehensive repository of biomedical research papers. We have previously used PubMed papers to pretrain state-of-the-art biomedical large language models (e.g., PubMedBERT~\citep{gu2021domain}, BioGPT~\citep{luo2022biogpt}). Here, we propose to further leverage the abundant figure-caption pairs in PMC full-text articles for vision-language pretraining. PMC contains 4.4 million publicly available full-text articles (as of June 15, 2022). We downloaded and extracted compressed directories with complete article packages.
Each article is represented as a package of XML, PDF, media, and supplementary materials. We extracted figure files and the matching captions, along with the PMID and the PMCID of the provenance articles. This yields a dataset with 15,282,336 image-caption pairs. 
We used Azure Databricks to process the data as it offered a scalable and reliable platform that is designed to handle large datasets and complex workflows in parallel using Apache Spark (\textbf{Fig. \ref{fig:overview}C}). 

We demonstrate the summary statistics of \ourdata{} in \textbf{Fig. \ref{fig:overview}A}. We found that biomedical images are much larger than the standard image size ($224\times224$) in general domains and biomedical image captions are much longer than the default max length (77) used by the standard CLIP method \citep{radford2021learning}. This necessitates the exploration of new model configurations for pretraining biomedical images and text. To probe the diversity and coverage of image classes in \ourdata{}, we created \ourdatalarge{} by splitting each scientific figure into individual panel images (see \textbf{Methods}). We then used a curated taxonomy \cite{GMB2015} with manually assigned image type keywords to estimate the frequency of each image type by assigning each image with the closest keywords in the learned embedding space from BiomedCLIP. The numbers likely overcount real class frequency but are close enough for a ball-park estimate. \textbf{Fig. \ref{fig:overview}B} shows the top 30 image types in \ourdatalarge{}. Images in PMC 
are extremely diverse, ranging from generic biomedical illustration (e.g., statistical figures, graphs, charts, tables and forms) to radiography (e.g., magnetic resonance, computerized tomography, and X-ray) to digital pathology and microscopy (e.g., light microscopy, and electron microscopy), among others. The large size and diversity raise our confidence that \ourdata{} can be used to train a state-of-the-art foundation model for biomedical vision-language processing.

\paragraph{Modeling}
\ourmodel{} is an advanced adaptation of the CLIP (Contrastive Language–Image Pretraining) model~\cite{radford2021learning}, specifically tailored for the biomedical domain. CLIP trains image and text encoders to embed $($image, text$)$ pairs in a shared space, optimizing for a high cosine similarity among positive pairs and a low similarity for negative pairs through the InfoNCE loss~\cite{oord2018representation}. Unlike the original CLIP, which was trained from scratch on internet-sourced image-text pairs, \ourmodel{} adapts this approach to better suit the unique characteristics of biomedical images and texts. This adaptation involves employing a domain-specific language model, PubMedBERT~\cite{gu2021domain}, in place of general-domain GPT-2~\cite{radford2019language} for the text encoder, and adjusting the tokenizer and context size to accommodate the typically longer biomedical literature. This contrasts with PubMedCLIP~\cite{eslami2021does} and MedCLIP~\cite{wang2022medclip}, where PubMedCLIP simply fine-tunes the original CLIP on a limited dataset from PubMed Central and MedCLIP incorporates medical knowledge into the learning process but still within the constraints of the original CLIP architecture. \ourmodel{} also introduces domain-specific adaptations to the image processing, utilizing larger Vision Transformer models~\cite{dosovitskiy2020image} and higher image resolutions to better capture the detailed visual information necessary for biomedical understanding. Additionally, it implements a patch dropout strategy~\cite{li2022scaling} to maintain pretraining efficiency while enhancing model performance. The tailored batch size optimization of \ourmodel{} further underscores its custom adaptation for the biomedical domain.

\paragraph{Benchmark} The overarching objective of pretraining is to improve performance across a broad spectrum of downstream applications. In general domains, comprehensive benchmarks like ELEVATER \citep{li2022elevater} have spurred rapid advances in vision-language pretraining by facilitating direct comparison among pretrained models.
In contrast, prior work on biomedical vision-language pretraining tends to use different tasks and datasets for downstream evaluation, e.g., \cite{zhang2020contrastive,wang2022medclip,eslami2021does}.
To facilitate evaluations of biomedical vision-language pretraining and expedite progress in biomedical vision-language processing, we compile a collection of eight standard vision-language datasets spanning key downstream tasks: image-to-text and text-to-image retrieval, image classification, and visual question answering (\textbf{Fig. \ref{fig:overview}D}). \textbf{Table \ref{tab:benchmark}} provides an overview of the datasets used in downstream applications.

\subsection*{\ourmodel{} enables accurate cross-modal retrieval}

\begin{figure}[]
    \centering
    \includegraphics[width=.95\textwidth]{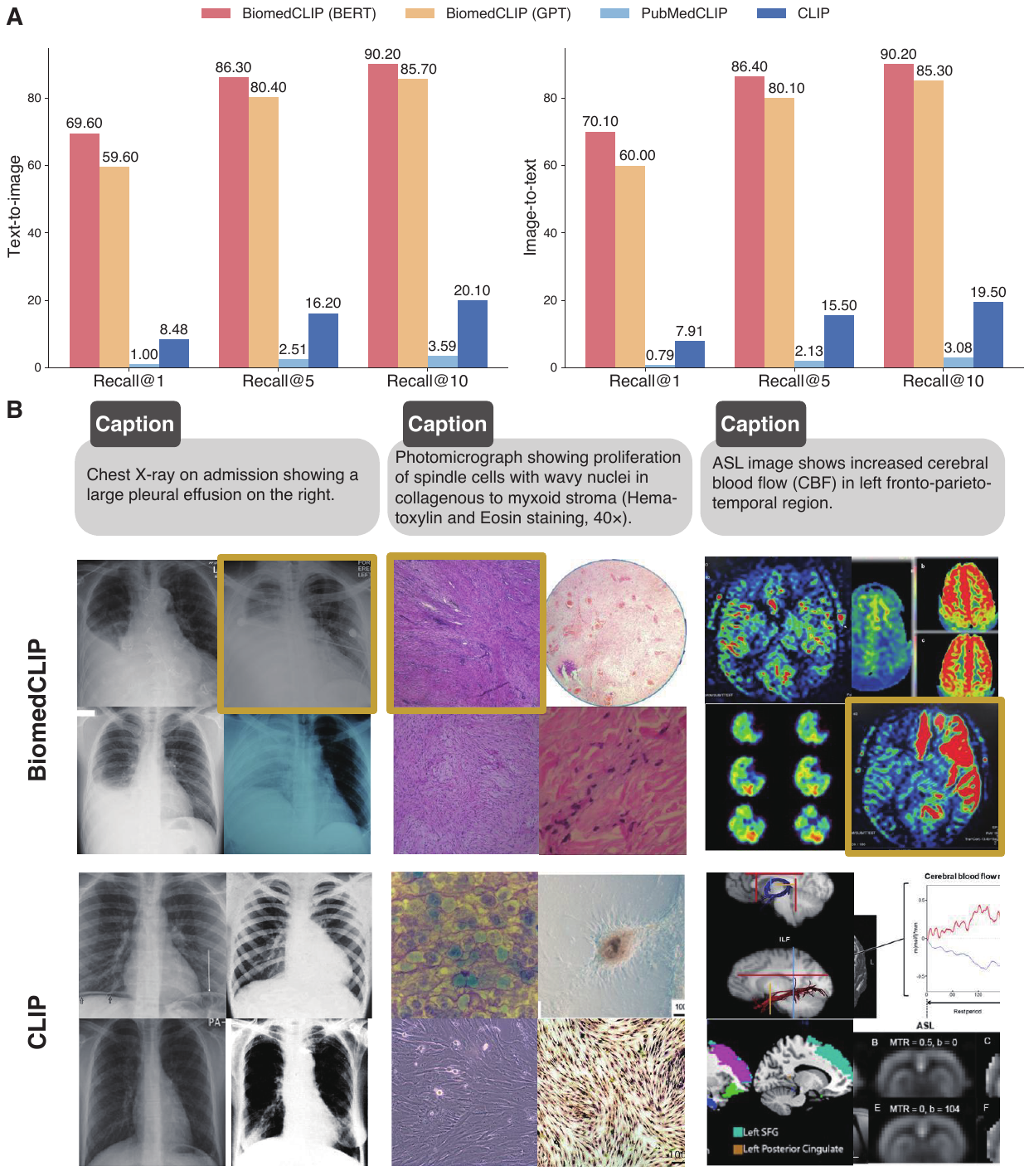}
    \caption{ \textbf{Comparison on cross-modal retrieval.} \textbf{A}: Test results on text-to-image retrieval (left) and image-to-text retrieval (right). X-axis shows different evaluation metrics, including Recall@1, Recall@5, and Recall@10. BiomedCLIP (BERT) and BiomedCLIP (GPT) denote using PubMedBERT and GPT as the encoder respectively. \textbf{B}: Three examples comparing \ourmodel{} and general-domain CLIP on text-to-image retrieval for sample PMC captions (top-4 predictions). Gold box indicates the ground truth figure for the caption.}
    \label{fig:retrieval}
\end{figure}

\noindent We first evaluate the task of cross-modal retrieval which aims to retrieve the corresponding image from the caption (text-to-image retrieval) or vice versa (image-to-text retrieval). The retrieval tasks mirror various image search and text generation in real-world applications and can be evaluated automatically in held-out image-text pairs. We use a held-out test set from \ourdata{} comprising 725,739 PMC image-caption pairs. The results are summarized in \textbf{Fig. \ref{fig:retrieval}A}. We first notice that the general-domain CLIP model performs poorly in the biomedical domain, necessitating the development of a domain-specific vision language model. In contrast, \ourmodel{} attains remarkably high retrieval accuracy: out of over 700 thousand candidates, \ourmodel{}'s top-5 results contain the correct one over 77\% of times, and its top-1 results are correct over 56\% of times.
To our surprise, PubMedCLIP \citep{eslami2021does} performs even worse than CLIP, despite biomedical adaptation. We attribute this to the training data that PubMedCLIP used. Despite being named after PubMed, PubMedCLIP only used a small set of radiology image-text pairs in the continual pretraining of CLIP, which accounts for a small fraction of images in the biomedical literature. Moreover, continual pretraining on a small dataset without extra care (e.g., augmenting with pretraining objective) may be prone to catastrophic forgetting \citep{mccloskey1989catastrophic}. 
In contrast, \ourmodel{} performs very well by pretraining on large-scale data from \ourdata{}, further indicating the importance of utilizing a diverse and large dataset for domain-specific vision language models.

To understand how \ourmodel{} outperforms general-domain CLIP in biomedical cross-modal retrieval, we show three random examples in \textbf{Fig. \ref{fig:retrieval}B}. In each example, we show the top-4 image retrieval results given the text prompt, with the correct answer shown in a gold box. General CLIP can find images matching common keywords such as ``chest X-ray", but have trouble differentiating subtle semantics such as ``pleural effusion", ``spindle shaped cells'', or even important biomedical image categories like ``ASL'' (Arterial Spin Labeling). By contrast, \ourmodel{} recognizes not only high-level categories but also details like ``a large pleural effusion on the right''. For example, the images retrieved by \ourmodel{} in the first example are completely correct according to the caption, except the right-bottom one, which still looks quite similar to the others. In the second example, \ourmodel{} is able to find the correct answer and additional H\&E images with spindle-shaped cells, as described in the prompt, unlike general CLIP. We attribute \ourmodel{}'s ability of capturing both coarse-grained and fine-grained semantics to our diverse and large pretraining dataset \ourdata{}.

\subsection*{\ourmodel{} enables accurate biomedical image classification}
\begin{figure}[]
    \centering
    \includegraphics[width=.95\textwidth]{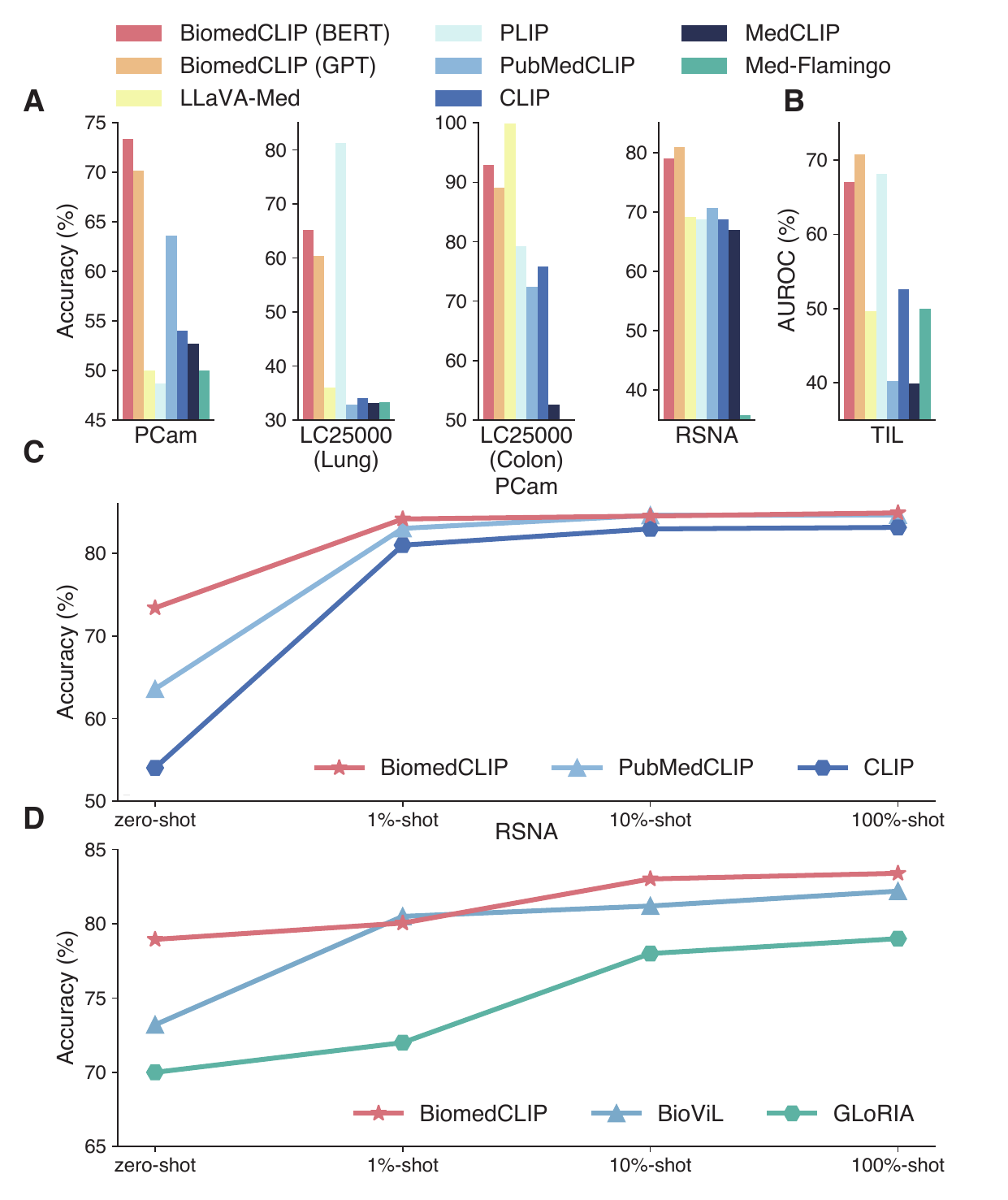}
    \caption{\textbf{Comparison on image classification.} \textbf{A,B}: Zero-shot image classification test results on five standard datasets. Following standard practice in prior work, we report test AUROC (\%) for TCGA-TIL, and accuracy (\%) for others. (TCGA-TIL is heavily class-imbalanced, unlike other datasets.) MedCLIP was evaluated on a sub-sample of RSNA in its original paper but is evaluated on the full dataset here for head-to-head comparison with other methods.  MedCLIP, PubMedCLIP and Med-Flamingo were pretrained on image-text pairs from biomedical research papers, whereas PLIP was pretrained on image-text pairs extracted from Twitter data. LLaVA-Med extends \ourmodel{} by fine-tuning it on instruction-following data for image-language conversations.
    \textbf{C,D}: Fine-tuned (linear probing) test accuracy on PCam (\textbf{C}) and RSNA (\textbf{D}).  GLoRIA and BioViL are state-of-the-art radiology-specific models tailored for radiology-specific tasks such as RSNA. Their results are taken from the respective papers.}
    \label{fig:img-class}
\end{figure}

We next evaluate the performance of \ourmodel{} for biomedical image classification on five datasets (see \textbf{Methods}). We investigate the zero-shot performance on \ourmodel{} models, where a short text prompt describing the class is used to assist the classification (\textbf{\Cref{tab:prompts}}). \ourmodel{} exhibits superior performance on zero-shot classification, attaining the highest overall accuracy (mean of the scores) across all the datasets (\textbf{Fig. \ref{fig:img-class}A,B}). PubMedCLIP was continual pretrained using radiology data so it outperforms the general-domain CLIP on the RSNA benchmark but generally yields inferior performance on the digital pathology benchmarks such as LC25000 and TCGA-TIL. 
PLIP~\cite{huang2023visual} was pretrained on digital pathology data from social media, so performs reasonably well on some of the pathology benchmarks, but yields much inferior performance on RSNA. Surprisingly, PLIP also performs rather poorly on the pathology benchmark PCam. We suspect that the pathology images in PCam (lymph nodes with metastatic tumors) might be under-represented in the social media data PLIP was pretrained on. Med-Flamingo \cite{moor2023med} was pre-trained as a few-shot learner on interleaved image-text data, with its pre-training data formulated as open-ended visual question answering (VQA). It almost always generates the same answer or choice regardless of the prompt template or test example, likely due to its pre-training nature, which limits its capability for zero-shot multimodal reasoning and hinders its ability to follow instructions for closed-set classification.
In contrast, \ourmodel{} achieves good results on all datasets from diverse image sources, again indicating the importance of pretraining on a diverse and large-scale dataset for building a foundational vision-language model. 

We also evaluate few-shot and full-shot performance on PCam and RSNA by linear probing the models with 1\%, 10\%, and 100\% of training data, respectively  (\textbf{Fig. \ref{fig:img-class}C,D}). Notably, by pretraining on diverse data across all biomedical image classes, \ourmodel{} even outperforms the state-of-the-art radiology-specific BioViL model \citep{boecking2022making} on the standard radiology benchmark RSNA. Furthermore, \ourmodel{} already outperforms fully supervised BioViL using only 10\% of labeled data. As shown in (\textbf{Fig. \ref{fig:overview}B}), the radiology-related images used in \ourmodel{} pretraining are no more than MIMIC-CXR used in BioViL pretraining, and the image-text pairs are likely to be much noisier. This excludes the possibility that the superior performance of \ourmodel{} on RSNA stems from more radiology-specific pretraining. Instead, the overall pretraining on the large and diverse \ourdata{}, including non-radiology image types, has helped pretrain a more robust image encoder, again indicating the superiority of using \ourdata{} to develop a multimodal foundation model.

\subsection*{\ourmodel{} improves medical visual question answering}
\begin{figure}[]
    \centering
    \includegraphics[width=.95\textwidth]{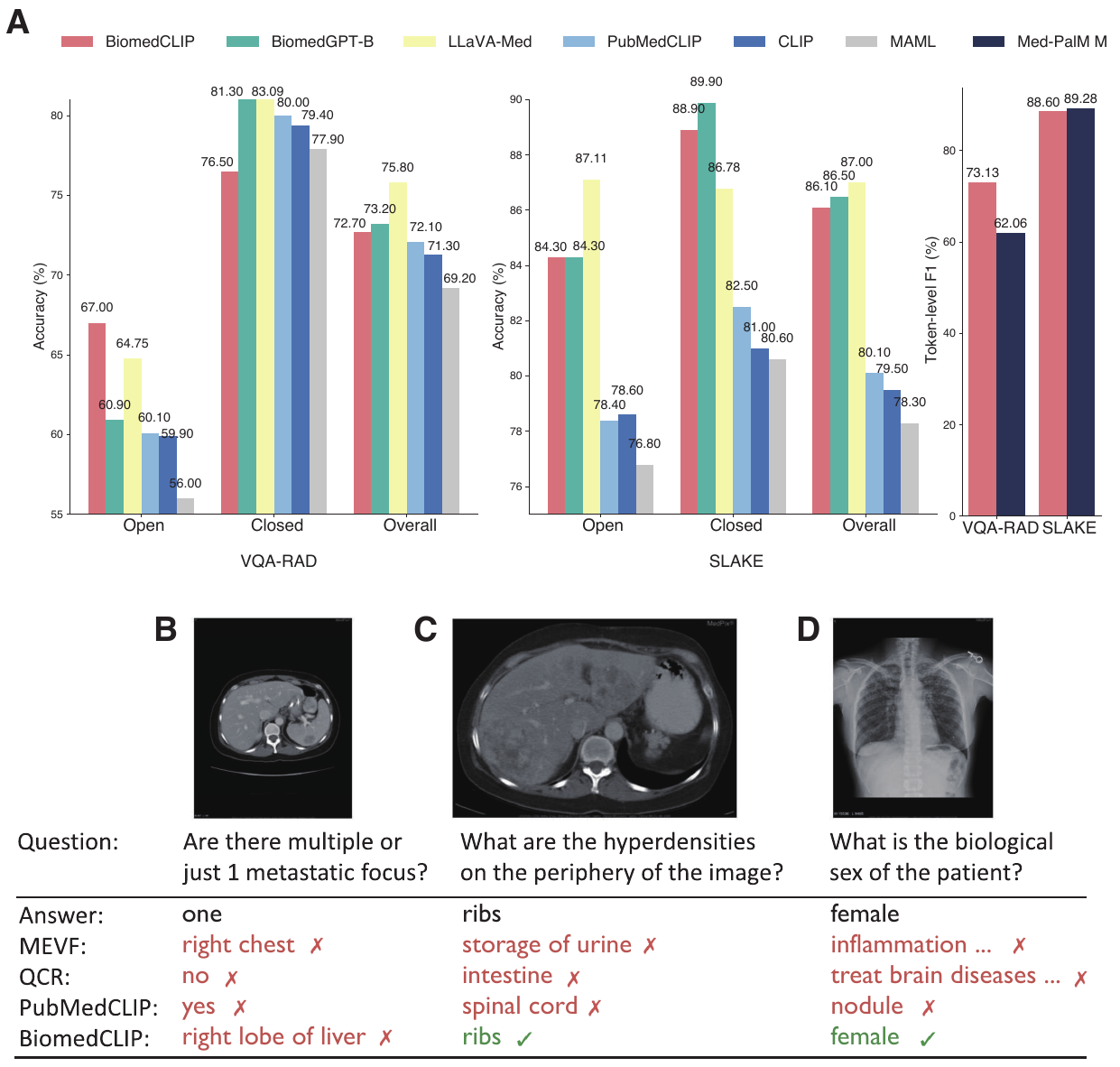}
    \caption{\textbf{Comparison on medical visual question answering.} \textbf{A}: Accuracy and token-level F1 on VQA-RAD and SLAKE under three different settings. Open: open-ended answers. Closed: multiple choice (mostly yes and no). \textbf{B}: Three examples from VQA-RAD as selected by the PubMedCLIP paper\cite{eslami2021does}, where all previous VQA models failed to produce the correct answer (including PubMedCLIP, the prior state of the art on this task). BiomedCLIP correctly answers the questions in C and D. While not technically correct, its answer to B nevertheless correctly identifies liver as the metastatic focus (on the right side of the CT scan).}
    \label{fig:vqa}
\end{figure}

After observing the superior performance of \ourmodel{} on image classification, we then evaluate it on medical visual question answering (VQA). Following the standard approach in prior work, we formulate VQA as a classification task using the METER \citep{dou2022meter} framework (see \textbf{Methods}). We evaluate the models on two standard datasets: VQ-RAD and SLAKE (\textbf{Fig. \ref{fig:vqa}A}). We report the respective accuracies for open-ended questions and closed-ended questions as well as the token-level F1. Again, \ourmodel{} exhibits superior performance compared to the radiology-specific state-of-the-art PubMedCLIP on both radiology benchmarks, with about one point and six points increases in overall accuracy respectively. \ourmodel{}'s gains are particularly pronounced for open-ended questions in VQA-RAD and for all question types in SLAKE. Furthermore, we see that \ourmodel{} achieves comparable performance comparing to the much larger model Med-PalM M (562B) \cite{tu2024towards} and the instruction-aware fine-tuned model BiomedGPT-B~\cite{zhang2023biomedgpt}. Notably, LLaVA-Med \cite{li2024llava}, which integrates \ourmodel{} as the core vision encoder into the LLaVA framework and further fine-tunes it to follow open-ended instructions, achieves the highest VQA performance on both datasets.

To further examine the answers provided by \ourmodel{}, we evaluate \ourmodel{} on some of the most challenging examples highlighted in the PubMedCLIP paper \citep{eslami2021does}, where all prior state-of-the-art models, including PubMedCLIP, failed to answer correctly (\textbf{Fig. \ref{fig:vqa}B-D}). In 
 the second example (\textbf{Fig. \ref{fig:vqa}C}), prior models fail to return the right answer, whereas in the third example (\textbf{Fig. \ref{fig:vqa}D}), their answers indicate that they even fail to understand what the question is about. \ourmodel{} nails both answers perfectly. In the first example, which is the most challenging one (\textbf{Fig. \ref{fig:vqa}B}), MEVF misidentifies the body part as displayed in the image, whereas QCR and PubMedCLIP misinterpret the question as a binary one (yes/no). While \ourmodel{} didn't get the correct answer either, it correctly identified the relevant organ as presented in the image. Since medical visual question answering requires both strong text understanding and image understanding ability, we attribute the superior results of \ourmodel{} to the large number of parallel image-caption pairs in \ourdata{}.


\subsection*{\ourmodel{} and \ourdata{} enable privacy-preserving analysis of proprietary data}
\begin{figure}[]
    \centering
    \includegraphics[width=.95\textwidth]{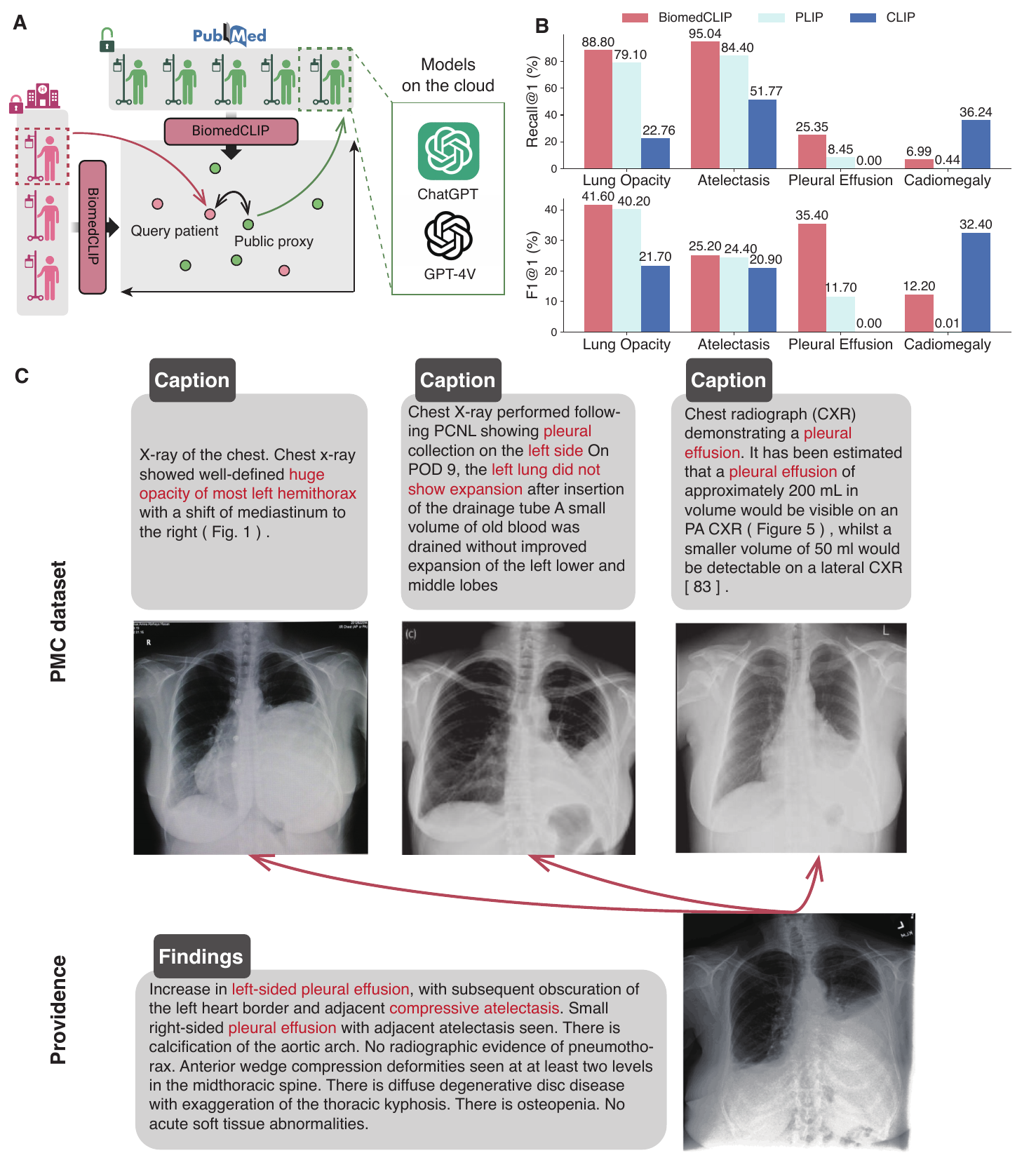}
    \caption{\textbf{Privacy-preserving proprietary data analysis.} \textbf{A}: \ourdata{} dataset is used as a proxy for proprietary patient data so that proprietary data can be analyzed using external models without being exposed to an external classifier. \textbf{B}: Comparing the classified labels between the proprietary Providence radiology data and the most similar \ourdata{} data in terms of Recall and F1. CheXbert is used as the external classifier to derive the labels. \textbf{C}: An example proprietary radiology image from Providence (bottom) and the top three most similar \ourdata{} images retrieved by \ourmodel{} (top). CheXbert is applied to the caption in \ourdata{} and the clinical report in Providence data to assess their similarity.}
    \label{fig:prov}
\end{figure}

Finally, we investigate whether \ourmodel{} and \ourdata{} can be used for privacy-preserving analysis of proprietary patient data (\textbf{Fig. \ref{fig:prov}A}). Proprietary patient data are often not allowed to be shared with external models, such as GPT4, preventing clinicians from using these models to analyze them. To tackle this, our idea is to use \ourdata{} as a proxy for proprietary patient data. In particular, for each proprietary data point, we will use \ourmodel{} to retrieve the most similar data points in \ourdata{}. We will then query external models using these similar public data points and aggregate the answers as the proxy output for the proprietary data point. During this process, the proprietary data point can be analyzed using external models without being exposed to GPT4.

To test the accuracy of this approximation, we collected $980$ radiology images and the associated clinical reports from Providence (see \textbf{Methods}) and treated them as proprietary patient data. We then compared the CheXbert-based labels assigned to the proprietary data and those labels assigned to the most similar data points in \ourdata{}. We observe a good agreement between these labels in terms of recall and F1, which is substantially higher than the agreements by PLIP and CLIP. In particular, our method can get 88.80 and 95.04 recall on lung opacity and atelectasis (\textbf{Fig. \ref{fig:prov}B}), indicating that \ourmodel{} can accurately retrieve relevant data points in \ourdata{} as high-quality proxies of proprietary radiology data. Meanwhile, \ourmodel{} gets a lower recall than CLIP on Cardiomegaly (\textbf{Fig. \ref{fig:prov}B}), suggesting that the approximation using \ourmodel{} could be limited by the captions in the scientific papers, which might not present all conditions as comprehensively as clinical reports. Finally, we illustrated an example proprietary image from Providence with the top three retrieved PMC images using \ourmodel{} (\textbf{Fig. \ref{fig:prov}C}), and observed that the captions of the retrieved PMC images could cover the main findings from the clinical report of the proprietary image, further raising our confidence that \ourmodel{} and \ourmodel{} can be high-quality proxies to enable the privacy-preserving analysis of proprietary data.

\section*{Discussion}
We present to our knowledge the largest study on biomedical vision-language pretraining using 15 million image-caption pairs extracted from PubMed Central full-text articles. 
Our pretraining data is at least two orders of magnitude larger than prior datasets, spanning an extremely diverse range of biomedical images. 
We conducted a systematic study on domain-specific adaptations for the biomedical domain and propose \ourmodel{} for biomedical vision-language processing. In extensive experiments on eight standard biomedical datasets, \ourmodel{} establishes new state of the art on standard applications such as cross-modal retrieval, image classification, and visual question answering. The promising results of our method indicate the effectiveness of \ourmodel{} on diverse biomedical image tasks and validate the advantage in conducting large-scale pretraining on extremely diverse data.

Our work is most relevant to biomedical multimodal representation learning. Most existing vision-language pretraining works focus on chest X-ray (CXR) with limited amounts of training data. ConVIRT \citep{zhang2020contrastive} pioneers the use of naturally occurring medical image-text pairs for self-supervision and demonstrates the potential of contrastive learning in vision-language pretraining.
Their image encoders benefit downstream CXR classification and retrieval tasks.
Their text encoder inherits a general-domain vocabulary, which leads to frequent encounters with out-of-vocabulary words when processing medical text.
While the word-piece tokenization mitigates this issue, common biomedical terms are often shattered into pieces, leading to suboptimal performance \citep{gu2021domain}.
GLoRIA \citep{huang2021gloria} uses the same general-domain vocabulary and 
extends ConVIRT by jointly learning multimodal global and local representations of medical images via contrasting attention weighted image regions with words in the paired reports.
LoVT \citep{muller2022joint} proposes a similar pretraining approach that aligns local representations of image regions and report sentences.
\cite{liao2021multimodal} learn multimodal representations by maximizing the mutual information between local features of medical images and text.
PubMedCLIP \citep{eslami2021does} fine-tunes the original CLIP on 80 thousand radiology image-caption pairs from the ROCO dataset \citep{pelka2018radiology}
 collected from PubMed Central \citep{roberts2001pubmed}, which is a tiny subset compared to \ourdata{}.
\cite{wang2021self} propose a transformer-based framework for mix-up image-text pretraining, which uses masked vision/language modeling for image-only or text-only data and uses binary cross entropy for paired image-text data.
They demonstrate the benefits of adopting pretrained models in three CXR applications, i.e., classification, retrieval, and image regeneration.
MedCLIP \citep{wang2022medclip} similarly extends contrastive learning to cover image-only and text-only data. They additionally introduce medical knowledge to alleviate false negatives.
MedAug \citep{pmlr-v149-vu21a} leverages patient metadata to select positive image pairs that go beyond augmentations of the same image.
BioViL \citep{boecking2022making} improves contrastive learning in self-supervised vision-language processing with radiology-specific semantic modeling tailored to CXR images and reports. It achieves the state of the art in radiology natural language inference and a range of CXR benchmarks. 
\cite{Iyer2022.11.19.22282519} show self-supervised multimodal pretraining on CXR data consistently outperforms ImageNet-pretrained models for CXR interpretation.

In the future, we would like to explore further improvement on pretraining and fine-tuning, multimodal generation, as well as real-world applications such as image search, digital pathology, and multimodal fusion for precision health. 
While \ourmodel{} shows clear benefits of large-scale domain-specific pretraining for biomedical vision-language processing, there are several limitations in our current method:
First, besides captions, in-line references (i.e., citances within the paper) can also be extracted and paired with the corresponding figures to create additional training signals. \ourdata{} currently doesn't include such data.
Second, half of the images in \ourdata{} are composite figures. 
Splitting such composite figures into sub-figures could enable more fine-grained modeling and potentially lead to better vision-language representations and grounding. 
\ourdatalarge{} is augmented with both in-line references and fine-grained image-text pairs. While our current us of \ourdatalarge{} is limited to tallying fine-grained image distribution, we plan to explore leveraging \ourdatalarge{} to enhance \ourmodel{} pretraining in future work. 
Third, due to computational constraints, the largest vision encoder we have explored is ViT-B, which is relatively small compared with ViT-L, ViT-H, and ViT-G. 
Similarly, we have been using a relatively low image resolution of 336. While this is more than sufficient for general-domain web images, biomedical images tend to have much higher resolution. E.g., more than 75\% of images in \ourdata{} have size larger than 336, as shown in \textbf{Fig. \ref{fig:overview}}. We plan to explore larger models and higher resolutions in future work.
Finally, it would be interesting to apply our methodology to additional biomedical modalities other than images, where similar
naturally co-occurring data abound, such as gene expression and sequence data along with textual descriptions.
\newpage
\section*{Methods}
\subsection*{Details of creating \ourdata{}}

PubMed Central Open Access Subset (PMC-OA)~\cite{PMC_Open_Access_Subset} contains 4.4 million publicly available full-text articles (as of June 15, 2022). 
We download PMC-OA from \href{https://www.ncbi.nlm.nih.gov/pmc/tools/ftp/\#indart}{ncbi.nlm.nih.gov/pmc/tools/ftp/\#indart} and extract complete packages for articles that include XML, PDF, media, and supplementary materials.
We use PubMed Parser~\cite{Achakulvisut2020} to parse the XML files and extract captions and the corresponding figure references, along with PMID and PMCID of the provenance articles.
This processing results in JSON objects stored in JSONL format, where each line represents one article, for convenient post-processing.
Articles without figure references, or with incorrectly formatted XML files with syntax errors or missing information errors are ignored.
After this clean-up, we collect 15 million figure-caption pairs (\ourdata{}) from over 3 million distinct articles.

\subsection*{Details of creating \ourdatalarge{}}

\ourdatalarge{} is created using a novel data curation pipeline that extracts and refines image-text pairs from \ourdata{}. The pipeline is designed to handle compound figures by breaking them down into sub-figures and captions, and it also incorporates in-line text references from articles as an additional data source for image-text pairs. The process begins with Data Ingestor, which downloads and processes article files. Then, Citance Extractor parses article text for figure references, and Caption Splitter uses regex and rules to separate figure captions into sub-captions with individual labels. Next, Citance Splitter assigns blocks of citances to each label, Optical Character Recognition (OCR) detects text in figures, and Label-To-Box Matcher matches labels with OCR-detected text. Finally, Figure Splitter segments compound images into panels (sub-figures), and Label-to-Panel Matcher assigns labels to the correct panels.
The pipeline overcomes various challenges like inconsistent labeling styles, OCR errors, and the ambiguous positioning of labels relative to sub-figures. Advanced techniques and heuristics are employed, such as string matching for OCR text, layout analysis to correct OCR mistakes, and region-based heuristics to match labels with panels. Through this intricate process, the \ourdatalarge{} dataset, with 46 million image-text pairs, is carefully constructed to facilitate the development of vision-language representations in the biomedical domain.

\subsection*{Cross-modal retrieval}
To evaluate the retrieval performance, we follow prior work \citep{schuhmann2022laion} to precompute image and text embeddings and perform an approximate nearest neighbor search.
Specifically, we use the pretrained vision encoder and text encoder from the CLIP models to precompute embeddings of figures and captions respectively. Given a figure embedding, we compute its cosine similarities with all captions in the test set and retrieve the $k$ most similar captions. Our evaluation metric measures if the original caption for the figure is within the $k$ retrieved captions, i.e., recall at top-$k$ or R@$k$. 
Similarly, we evaluate Recall@$k$ for text-to-image cross-modal retrieval.

We consider several CLIP models as baselines: OpenAI CLIP~\cite{radford2021learning}, which is pretrained on 400 million general-domain (image, text) pairs collected from the Internet; PubMedCLIP~\cite{EslamiDeMeloMeinel2021CLIPMedical}, which fine-tunes OpenAI CLIP on 80k radiology image-caption pairs. 
We also compare with a variant of our \ourmodel{} (i.e., BiomedCLIP ViT-B/16-224-GPT/77), which continues pretraining OpenAI CLIP on \ourdata{}.

\subsection*{Biomedical image classification}
We use the evaluation toolkit ELEVATER \citep{li2022elevater} to facilitate our experiments on image classification. ELEVATER is an easy-to-use toolkit that can efficiently adapt pretrained vision-language models and automatically tune hyper-parameters. It supports zero-shot, few-shot and full-shot evaluations, with linear probing and full model fine-tuning available for the latter two settings. It also contains twenty image classification datasets collected from various domains, including a biomedical one PatchCamelyon, which we use in our experiments. In addition, we evaluate on three standard biomedical imaging benchmarks LC25000, TCGA-TIL and RSNA. 

An overview of the datasets can be found in Table \ref{tab:benchmark}. \textbf{PatchCamelyon} (PCam) \citep{Veeling2018-qh} contains $327,680$ color images (96$\times$96px), which were taken from histophathology scans of lymph node sections. The images have been assigned  a binary label indicating whether or not they contain metastatic tissue.
\textbf{LC25000} \citep{borkowski2019lung} contains $25,000$ histopathology images (768$\times$768px). These images were generated by augmentation from a collection of HIPAA-compliant, validated sources originally comprising 750 images of lung tissue (250 benign, 250 adenocarcinomas, and 250 squamous cell carcinomas) and 500 images of colon tissue (250 benign and 250 adenocarcinomas). The dataset is divided into five classes: lung benign tissue, lung adenocarcinoma, lung squamous cell carcinoma, colon adenocarcinoma, and colon benign tissue, with each class containing 5,000 images.
\textbf{TCGA-TIL} \citep{saltz2018tumor, saltz2018spatial} contains $2,480$ image patches (500$\times$500px) that were partitioned from H\&E whole-slide images from The Cancer Genome Atlas (TCGA) \citep{clark2013cancer} lung adenocarcinoma (LUAD) cases (5.9\% of the patches are labelled as LUAD).
\textbf{RSNA} Pneumonia \citep{shih2019augmenting} contains about $30,000$ frontal-view chest radiographs collected from the National Institutes of Health's public database of chest X-rays. It contains binary labels classifying pneumonia against normal cases. 

We compare our method against competing methods, including CLIP, MedCLIP, PubMedCLIP, PLIP, Med-Flamingo and LLaVA-Med. MedCLIP \citep{wang2022medclip} extends the pretraining to include large unpaired images and texts through contrastive learning. It uses the pretrained BioClinicalBERT and Swin Transformer \citep{liu2021Swin} as the backbone text encoder and visual encoder respectively, and fine-tunes on MIMIC-CXR and CheXpert datasets. PubMedCLIP \citep{eslami2021does} fine-tunes CLIP on the  Radiology Objects in COntext (ROCO) dataset \cite{pelka2018radiology}, which consists of 80K radiology image-text pairs drawn from PubMed articles.
PLIP~\cite{huang2023visual} was pretrained on OpenPath, which contains about 208K pathology images paired with natural language descriptions from medical Twitter.
Med-Flamingo~\cite{moor2023med} is a multimodal few-shot learner based on OpenFlamingo-9B, further pre-trained with 2M interleaved medical image-text data from publications and textbooks, formulated for
open-ended visual question answering. 
LLaVA-Med~\cite{li2024llava} extends \ourmodel{} and is further tuned to master open-ended conversational
semantics using GPT-4 generated instruction-following data.
All the models, except the last two which use prompts to answer close-ended questions, are adapted into ELEVATER for evaluation. 

\subsection*{Medical Visual Question Answering}
We utilize the METER \citep{dou2022meter} framework to facilitate our experiments on visual question answering (VQA). It formulates the VQA task as a classification task. The core module of METER is a transformer-based \textit{co-attention} multimodal fusion module that produces cross-modal representations over the image and text encodings, which are then fed to a classifier for predicting the final answer. We compare \ourmodel{} with general-domain CLIP, MAML (Model-Agnostic Meta-Learning) network pretrained only on visual data and PubMedCLIP. All three models were fine-tuned for VQA tasks using the QCR (Question answering via Conditional Reasoning) framework \citep{zhan2020medical} that alternatively uses a MLP-based attention networks with conditional reasoning as the fusion module. Results for MAML, CLIP and PubMedCLIP are collected from \cite{eslami2021does}. 
We also compare our method with instruction-aware fine-tuned models BiomedGPT-B \cite{zhang2023biomedgpt}, which was fine-tuned on 25 curated datasets, LLaVA-Med and significantly larger model Med-PaLM M (562B)~\cite{tu2024towards}, which was fine-tuned on 12
open source datasets and 14 individual tasks.


We evaluated our method and competing methods on VQA-RAD~\citep{lau2018dataset} and SLAKE~\citep{liu2021slake} using accuracy and token-level F1 metrics. The accuracy metric evaluates string-level correctness by treating VQA as a classification task with a set of predefined answer candidates collected from the training data, while the token-level F1 metric assesses token-level matching by feeding the answers into GPT-2 to obtain tokens. The dataset VQA-RAD consists of $315$ radiology images and $3,515$ question-answer pairs that were manually constructed by clinicians. Images in the test set are also present in the training set but the question-answer pairs do not overlap. SLAKE (English only)  consists of $642$ radiology images and over $7,000$ question-answer pairs annotated by experienced physicians. It covers more human body parts than VQA-RAD and does not have common images between the training and test sets. See \Cref{tab:benchmark} for details. 

\subsection*{Privacy-preserving proprietary data analysis}
We collected 980 chest X-ray images and the associated clinical reports from Providence Healthcare organization and treated them as proprietary data. We employed \ourmodel{} for image-to-image retrieval to find the most similar images in the PMC-15M dataset. To optimize computational efficiency, we downsampled \ourdata{} in two steps. Initially, we sampled $1,000$ MIMIC-CXR seed images and used \ourmodel{} to retrieve $100,000$ CXR-relevant images from \ourdata{}. Subsequently, we utilized \ourmodel{} to filter out the 90\% least consistent image-caption pairs, based on low similarity scores between image and caption embeddings, resulting in a refined subset of $10,000$ CXR images from \ourdata{} used for final image-to-image retrieval.

We obtained 13 CheXbert labels, excluding `No Findings', for the proprietary images by running CheXbert on the Findings and Impression sections of their corresponding clinical reports. Each label was classified as either \textit{positive} or \textit{negative}, with \textit{uncertain}, \textit{blank}, and \textit{negative} outputs from CheXbert all categorized as \textit{negative}. For the retrieved PMC images, we acquired labels by applying CheXbert to their corresponding captions. We conducted similar image-to-image retrieval experiments using PLIP and CLIP, evaluating label agreement between the top one retrieved and proprietary images through F1 and Recall metrics.

\section*{Data availability}
We will provide scripts to reproduce \ourdata{} and \ourdatalarge{} from PubMed Central Open-Access Data (PMC-OA), upon publication of this manuscript. These scripts will be made available at \url{https://aka.ms/biomedclip}. (Currently, this url is pointed to a preliminary model release at Hugging Face.)

\section*{Code availability}
\ourmodel{} will be made fully available at \url{https://aka.ms/biomedclip}, including the model weights and relevant source code for pretraining, fine-tuning, and inference. We will also provide detailed methods and implementation steps to facilitate independent replication.


\newpage
\bibliography{references}
\bibliographystyle{naturemag}

\newpage
\begin{table}[!ht]
\centering
\begin{tabular}{>{\centering}m{0.14\textwidth}>{\centering}m{0.1\textwidth}>{\centering}m{0.1\textwidth}>{\raggedright}m{0.3\textwidth}ccc}

\multirow{2}{*}{\textbf{Task}} & \multirow{2}{*}{\textbf{Dataset}} & \multirow{2}{*}{\textbf{Metric}} & \multirow{2}{*}{\textbf{Description}} & \multicolumn{3}{c}{\textbf{Data size}} \\ 
\cline{5-7}
 &  &  &  & Train & Dev & Test \\ 
\shline
Cross-Modal Retrieval & \ourdata{} & Recall@k & Given textual description (caption), retrieve the corresponding image, or vice versa. (Image size: see \textbf{Fig. \ref{fig:overview}A}) & 13.9M & 13.6k & 726k \\ \hline
& PCam & Accuracy & Binary classification on whether a histopathology image of lymph node contains metastatic tumor tissue. (Image size: 96$\times$96) & 262,144 & 32,768 & 32,768  \\
\cline{2-7}
& LC25000 (Lung) & Accuracy & Ternary classification (benign, adenocarcinoma, squamous cell carcinoma) on histopathology images of lung tissue. (Image size: 768$\times$768) & - & - & 15,000 \\
\cline{2-7}
Image\quad Classification & LC25000 (Colon) & Accuracy & Binary classification (benign, adenocarcinoma) on histopathology images of colon tissue. (Image size: 768$\times$768)  & - & - & 10,000 \\ 
\cline{2-7}
& TCGA-TIL & AUROC & Binary classification on whether lung H\&E whole-slide image patches show adenocarcinoma. (Image size: 512$\times$512) & - & - & 2,480 \\ 
\cline{2-7}
 & RSNA & Accuracy &  Binary classification on whether chest X-rays show pneumonia. (Image size: 500$\times$500) & 18,678 & 4,003 &  9,069 \\ 
\hline
\multirow{4}{*}{VQA} & VQA-RAD & Accuracy & Answer clinician questions about radiology images. & 3,064 & - & 451 \\ 
\cline{2-7}
 & SLAKE & Accuracy & Answer clinician questions about radiology images (X-rays, and single slices of CTs and MRIs). & 4,919 & 1,053 & 1,061 \\
\shline
\end{tabular}
\caption{Tasks, datasets, and evaluation metrics used in our biomedical vision-language processing study. The numbers in train, dev, and test represent numbers of image-caption pairs in cross-modal retrieval, and numbers of images in image classification and visual question-answering (VQA).}
\label{tab:benchmark}
\end{table}

\begin{figure}[]
    \centering
    \includegraphics[width=.95\textwidth]{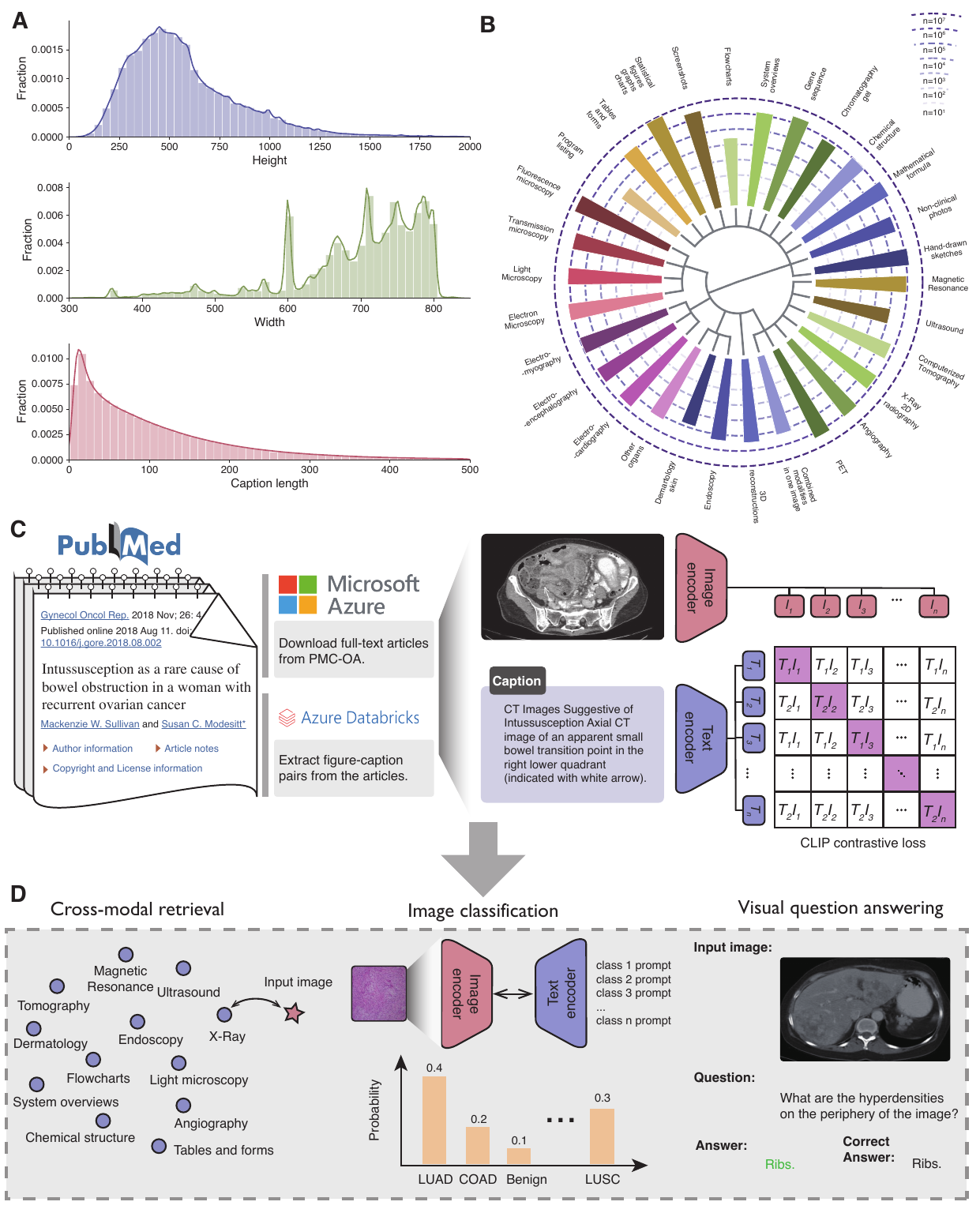}
    \caption{ \textbf{Overview of \ourmodel{} and \ourdata{}.} \textbf{A}: Statistics of image sizes and caption lengths in \ourdata{}. \textbf{B}: Distribution of image types. Image types are defined according to the image taxonomy in \cite{GMB2015}. \textbf{C}: BiomedCLIP collects 4.4 million publicly available full-text scientific articles. These articles are processed by Azure Databricks to obtain parallel image-text pairs. These pairs are then used to train a multimodal biomedical foundation model based on contrastive learning. \textbf{D}: BiomedCLIP excels on a variety of downstream applications such as cross-modal retrieval, zero-shot image classification, and visual question answering.}
    \label{fig:overview}
\end{figure}

\newpage
\newpage
\section*{Supplementary Note}
\subsection*{Details of \ourmodel{} architecture and ablation studies} 
\paragraph{Review of CLIP} We first give a brief review of the CLIP pretraining approach \citep{radford2021learning}.
Given a batch of $N$ (image, text) pairs, 
CLIP learns a multimodal embedding space by jointly training an image encoder and a text encoder to maximize the cosine similarity between the image and text embeddings of the $N$ pairs in the batch while minimizing the cosine similarity of the embeddings of the other $N^2-N$ non-pairs. Concretely, CLIP minimizes the InfoNCE loss \citep{oord2018representation}, i.e., a symmetric cross entropy loss over these similarity scores:
\begin{equation}
    \mathcal{L}=-\frac{1}{2N}(\sum_{i=1}^N\log\frac{e^{\text{cos}(\mI_i,\mT_i)/\tau}}{\sum_{j=1}^Ne^{\text{cos}(\mI_i,\mT_j)/\tau}} + \sum_{i=1}^N\log\frac{e^{\text{cos}(\mI_i,\mT_i)/\tau}}{\sum_{j=1}^Ne^{\text{cos}(\mI_j,\mT_i)/\tau}}),
\end{equation}
where $\tau$ is a learnable temperature parameter, directly optimized during training as a log-parameterized multiplicative scalar; 
$\mI_i$ and $\mT_i$ are embeddings for the $i$-th image and text, produced by a linear projection layer on top of the image encoder and text encoder.
Rather than initializing with pretrained weights, CLIP trains the image encoder and text encoder from scratch. For the image encoder, CLIP considers two different architectures, ResNet-50 \citep{he2016deep} and Vision Transformer \citep[ViT;][]{dosovitskiy2020image}. The text encoder is effectively GPT-2~\citep{radford2019language} based on transformer~\citep{vaswani2017attention}.

\paragraph{Adapting CLIP to \ourmodel{}} Biomedical text and images are drastically different from the web data used in CLIP pretraining. We find that the standard CLIP settings are suboptimal for biomedical vision-language pretraining. We thus conducted a systematic study of potential adaptations and identified a series of domain-specific adaptations for the biomedical domain. We used the optimization loss and cross-modal retrieval results on the validation set to guide our initial exploration and conducted detailed ablation studies.

On the text side, we replace a blank-slate GPT-2 with a pretrained language model more suited for biomedicine. 
Specifically, we initialize with PubMedBERT, which shows substantial gains from domain-specific pretraining~\citep{gu2021domain}.
Correspondingly, for the tokenizer, we replace Byte-Pair Encoding \citep[BPE;][]{sennrich-etal-2016-neural} with WordPiece \citep{kudo-richardson-2018-sentencepiece}, which uses unigram-based likelihood rather than shattering all words to characters and greedily forming larger tokens based on frequency. 
The original CLIP uses a context of 77 tokens, but biomedical text is typically longer, as shown in \textbf{Fig. \ref{fig:overview}A}. 
We thus increase the context size to 256, which covers 90\% of PMC captions.
\Cref{tab:txt-side} shows that both modifications bring substantial improvements over the original CLIP model on the validation set.

\setcounter{table}{0}
\renewcommand*{\tablename}{Supplementary Table}

\begin{table}[!ht]
\centering
\begin{tabular}{lccccc}
& & & & img2txt (\%) & txt2img (\%) \\ 
text encoder & vocab & context length & loss ($\downarrow$) & R@1($\uparrow$) & R@1($\uparrow$) \\ \toprule
GPT & 50k general domain & 77 & 0.6626 & 64.53 & 63.56 \\ 
PubMedBERT & 30k domain specific & 77 & 0.5776 & 69.03 & 67.41 \\ 
\rowcolor{defaultcolor} PubMedBERT & 30k domain specific & 256 & \textbf{0.4807} & \textbf{73.50} & \textbf{72.26} \\ 
\end{tabular}
\caption{Improvements from text-side domain-specific adaptations, as measured on the \ourdata{} validation set.
The training epochs are 8. All other hyperparameters are reported in \Cref{tab:hyperparameters}.
}
\label{tab:txt-side}
\end{table}

On the image side,
we first evaluated Vision Transformer (ViT) across different scales, ranging from ViT-Small, ViT-Medium, to ViT-Base. The suffix ``/16'' in the ViT model names refers to the patch size of 16$\times$16 pixels i.e., the input images are divided into patches of this size, and fed through the transformer blocks. As shown in \Cref{tab:vit-scales}, we found that larger ViT results in better performance, confirming the importance of model scalability on our new dataset \ourdata{}. 
We used the largest one (ViT-B/16) in all subsequent experiments.

\begin{table}[!ht]
\centering
\tablestyle{5pt}{1.1}
\begin{tabular}{cccccc}
& & & & img2txt (\%) & txt2img (\%) \\
vision encoder & trainable params & hidden dim & loss ($\downarrow$) & R@1($\uparrow$) & R@1($\uparrow$) \\ \toprule
ViT-S/16 & 22M & 384 & 0.5342 & 69.45 & 68.02 \\ 
ViT-M/16 & 39M & 512 & 0.5063 & 71.85 & 70.22 \\ 
\rowcolor{defaultcolor} ViT-B/16 & 86M & 768 & \textbf{0.4807} & \textbf{73.50} & \textbf{72.26} \\ 
\end{tabular}
\caption{Validation performance for various ViT models (Small, Medium, Base). All experiments use PubMedBERT to initialize the text encoder with the maximal context length of 256.
The training epochs are 8. All other hyperparameters are reported in \Cref{tab:hyperparameters}.
}
\label{tab:vit-scales}
\end{table}

\begin{table}[!ht]
\centering
\tablestyle{5pt}{1.1}
\begin{tabular}{clccc}
& & & img2txt (\%) & txt2img (\%) \\ 
vision encoder & initialization & loss ($\downarrow$) & R@1($\uparrow$) & R@1($\uparrow$) \\ \toprule
ViT-B/16 & random initialization & 0.3814 & 83.15 & 81.75 \\ 
\rowcolor{defaultcolor} ViT-B/16 & pretrained on ImageNet & 0.3819 & 82.90 & 81.86 \\ 
\end{tabular}
\caption{Validation performance for vision encoders initialized with different weights. All experiments use PubMedBERT to initialize the text encoder, with the maximal context length of 256.  All other hyperparameters are reported in \Cref{tab:hyperparameters}.}
\label{tab:img-side-pretrain-vs-scratch}
\end{table}

Next, we compared two different ways to initialize the vision encoder. \Cref{tab:img-side-pretrain-vs-scratch} shows that the vision encoder pretrained on ImageNet \citep{dosovitskiy2020image} does not have advantages over random initialization.   
However, in our downstream tasks, ImageNet-pretrained weights offer more stable performance. 
Therefore, we chose to initialize ViT-B/16 with ImageNet-pretrained weights.
Lastly, biomedical image understanding often requires fine-grained visual features~\citep{zhang2020contrastive}.
In \Cref{tab:img-side}, we compared two choices of input image resolution: 224$\times$224 and 384$\times$384.
By increasing image resolution, we observe significant gains in validation results. But this also leads to a doubling of pretraining time.
In addition, increased image resolution does not consistently enhance performance in downstream tasks. As Table \Cref{tab:zeroshot_sup} shows, \ourmodel{} exhibits inferior performance in zero-shot classification across all five datasets when using larger image size of 384 compared to 224. This discrepancy is particularly notable in PCam, where the raw image resolution (96$\times$96) is considerably smaller than the model's input image size. Upsampling images here may introduce noise, potentially contributing to the observed decrease in performance. Consequently, we opt for an image size of 224 in the subsequent experiments.

\begin{table}[!ht]
\centering
\tablestyle{5pt}{1.1}
\begin{tabular}{lcccc}
& & & img2txt (\%) & txt2img (\%) \\ 
image size & training time & loss ($\downarrow$) & R@1($\uparrow$) & R@1($\uparrow$) \\ \toprule
\rowcolor{defaultcolor} 224px & 1.00x & 0.3819 & 82.90 & 81.86 \\ 
384px & 1.92x & 0.3406 & 84.63 & 83.56 \\ 
\end{tabular}
\caption{Pretraining time and validation performance for vision encoders with different image sizes. All experiments use ViT-B/16 as the image encoder and PubMedBERT to initialize the text encoder, with a maximal context length of 256. All other hyperparameters are reported in \Cref{tab:hyperparameters}.}
\label{tab:img-side}
\end{table}

\begin{table}[!ht]
\centering
\tablestyle{5pt}{1.2}
\begin{tabular}{l|ccccc|c}
 & & LC25000 & LC25000 & TCGA- & &\\
image size & PCam & (Lung) & (Colon) & TIL & RSNA & mean \\
 \shline
\rowcolor{defaultcolor} 224px & \textbf{73.41} & \textbf{65.23} & \textbf{92.98} & \textbf{67.04} & \textbf{78.95} & \textbf{75.52} \\
384px & 67.15 & 61.80 & 87.42 & 57.00 & 78.49 & 70.37 \\
\end{tabular}
\caption{Performance in downstream zero-shot image classification for vision encoders with different image sizes. AUROC (\%) for TCGA-TIL; accuracy (\%) for others. All hyperparameters are reported in \Cref{tab:hyperparameters}.}
\label{tab:zeroshot_sup}
\end{table}


\begin{table}[!ht]
\centering
\tablestyle{5pt}{1.2}
\begin{tabular}{lcc}
& image-to-text retrieval (\%) & text-to-image retrieval (\%) \\ 
batch size & R@1($\uparrow$) & R@1($\uparrow$) \\ \toprule
2k & 79.69 & 78.43 \\ 
\rowcolor{defaultcolor} 4k & \textbf{82.90} & \textbf{81.86} \\ 
\end{tabular}
\caption{Validation performance with different batch size.}
\label{tab:batch-size2}
\end{table}

\begin{table}[!ht]
\centering
\tablestyle{5pt}{1.2}
\begin{tabular}{l|cc}
& img2txt (\%) & txt2img (\%) \\ 
batch size & R@1($\uparrow$) & R@1($\uparrow$) \\ \toprule
4k$\rightarrow$ 4k & 83.98 & 82.71 \\ 
\rowcolor{defaultcolor} 4k$\rightarrow$ 64k & \textbf{87.32} & \textbf{86.66} \\ 
\end{tabular}
\caption{Validation performance with constant batch size of 4k for all 40 epochs vs increasing batch size from 4k (in the first 8 epochs) to 64k (in the remaining 32 epochs).}
\label{tab:batch-size}
\end{table}

Finally, we investigated the impact of batch size.
In \Cref{tab:batch-size2}, we show larger batch size generally has better validation performance. 
In \Cref{tab:batch-size}, we studied increasing the batch size up to 64k to match the choices of \cite{radford2021learning,cherti2022reproducible}.
While there was a further increase in validation performance, we found that the gain did not translate to the downstream evaluation after reaching the batch size of 4k (\Cref{tab:batch-size2}). The potential explanation for this could be that an extremely large batch size requires more training data and longer epochs. CLIP~\citep{radford2021learning} uses 400M image-text pairs, while \ourdata{} has 15M pairs. We chose the 4k batch size to train our \ourmodel{}.

\paragraph{Putting it all together}
We pretrained a series of \ourmodel{} models on \ourdata{} using the optimal batch schedule above and compared them with general-domain CLIP models~\citep{radford2021learning}.
As \Cref{tab:final-set} shows, large-scale pretraining or continual pretraining on \ourdata{} is always helpful, and the best validation performance are generally attained using the biomedical pretrained language model (PubMedBERT), a larger vision transformer, and a higher image resolution.  All hyperparameters are summarized in \Cref{tab:hyperparameters}.

\begin{table}[!ht]
\centering
\tablestyle{4pt}{1.2}
\begin{tabular}{llccc}
& & & img2txt (\%) & txt2img (\%) \\
model & config & data & R@1($\uparrow$) & R@1($\uparrow$) \\ \toprule
OpenAI CLIP & ResNet-50-224-GPT/77 & WIT-400M & 10.31 & 10.38 \\ 
OpenAI CLIP & ViT-B/16-224-GPT/77 & WIT-400M & 11.82 & 11.65 \\ \midrule
\ourmodel{} & ResNet-50-224-GPT/77 & WIT-400M $\rightarrow$ \ourdata{}  & 81.17 & 80.17 \\ 
\ourmodel{} & ViT-B/16-224-GPT/77 & WIT-400M $\rightarrow$ \ourdata{}  & 81.57 & 80.89  \\ 
\ourmodel{} & ViT-B/16-224-BERT/256 & ImageNet/PubMed $\rightarrow$ \ourdata{} & 82.90 & 81.86 \\ 

\end{tabular}
\caption{Comparison of \ourmodel{} and general-domain CLIP models on validation performance. ``WIT-400M $\rightarrow$ \ourdata{}" indicates initialization with OpenAI CLIP weights that were pretrained on WIT-400M \citep{radford2021learning}, followed by continual pretraining on \ourdata{}. ``PMB/256" denotes PubMedBERT with the maximal context length of 256.}
\label{tab:final-set}
\end{table}

\begin{table}[!ht]
\centering
\tablestyle{5pt}{1.2}
\begin{tabular}{l|l}
\textbf{Hyperparameters} & \textbf{Value} \\ \shline
optimizer & AdamW \citep{loshchilov2017decoupled} \\
peak learning rate &  5.0e-4 \\
weight decay &  0.2 \\
optimizer momentum &  $\beta_1$, $\beta_2$ = 0.9, 0.98\\
eps & 1.0e-6 \\
learning rate schedule & cosine decay \\
epochs & 32 \\
warmup (in steps) & 2000 \\
random seed & 0 \\
image mean & (0.48145466, 0.4578275, 0.40821073) \\
image std  & (0.26862954, 0.26130258, 0.27577711) \\
augmentation & RandomResizedCrop \\
validation frequency & every epoch 
\end{tabular}
\caption{Hyperparameters for pretraining settings.}
\label{tab:hyperparameters}
\end{table}

\paragraph{Implementation}
Our implementation is based on OpenCLIP \citep{ilharco_gabriel_2021_5143773}, an open source software adapted for large-scale distributed training with contrastive image-text supervision. 
The pretraining experiments were conducted with up to 16 NVIDIA A100 GPUs or 16 NVIDIA V100 GPUs, via PyTorch DDP \citep{li2020pytorch,paszke2019pytorch}. To reduce memory consumption, we enable gradient checkpointing and automatic mixed precision (AMP) with datatype of bfloat16 (whenever supported by the hardware).
In addition, we use a sharding contrastive loss \citep{cherti2022reproducible}, which achieves identical gradients to InfoNCE \citep{oord2018representation} and reduces memory usage by eliminating redundant computations and only computing the similarities of locally relevant features on each GPU. 

\begin{table}[!ht]
\centering
\tablestyle{5pt}{1.2}
\begin{tabular}{l|ll}
 & \multicolumn{2}{c}{\ourmodel{}} \\
 dataset & \multicolumn{1}{c}{classes} & \multicolumn{1}{c}{templates} \\
 \shline
PCam & {\small \texttt{normal lymph node}} & {\small \texttt{this is an image of \{\}}}; \\
 & {\small\texttt{lymph node metastasis}} & {\small \texttt{\{\} presented in image}}\\\hline
LC25000 (Lung) &  {\small \texttt{lung adenocarcinomas}} & {\small \texttt{this is an image of \{\}}}; \\
 & {\small\texttt{normal lung tissue}} & {\small \texttt{\{\} presented in image}}\\
 & {\small \texttt{lung squamous cell carcinomas}} & \\\hline
LC25000 (Colon) &  {\small \texttt{colon adenocarcinomas}} & {\small \texttt{a photo of \{\}}}; \\
 & {\small\texttt{normal colonic tissue}} & {\small \texttt{\{\} presented in image}}\\\hline
TCGA-TIL & {\small \texttt{none}} & {\small \texttt{a photo of \{\}}}; \\
 & {\small\texttt{tumor infiltrating lymphocytes}} & {\small \texttt{\{\} presented in image}} \\\hline
RSNA & {\small \texttt{normal lung}} & {\small \texttt{a photo of \{\}}}; \\
 & {\small\texttt{pneumonia}} & {\small \texttt{\{\} presented in image}}\\
\end{tabular}
\caption{Prompts used for zero-shot image classification.}
\label{tab:prompts}
\end{table}

\end{document}